\documentclass[conference]{IEEEtran}
\IEEEoverridecommandlockouts
\usepackage{cite}
\usepackage{amsmath,amssymb,amsfonts}
\usepackage{algorithmic}
\usepackage{graphicx}
\usepackage{textcomp}
\usepackage{xcolor}

\newcommand{\circcirc}{\ensuremath{\circ\!\!-\!\!\circ}}
\newcommand{\circrightarrow}{\ensuremath{\circ\!\!\!\rightarrow}}

\newcommand{\circline}{\ensuremath{\circ\!-}}
\usepackage{algorithm}
\usepackage{algorithmic}
\usepackage{subfig}
\usepackage{tikz}
\usetikzlibrary{arrows.meta}
\tikzset{%
  >={Latex[width=2mm,length=2mm]},
            base/.style = {rectangle, rounded corners, draw=black,
                           minimum width=2cm, minimum height=1cm,
                           text centered, font=\sffamily},
  activityStarts/.style = {base, minimum width=1cm, fill=orange!15, font=\ttfamily},
       startstop/.style = {base, minimum width=1cm, fill=orange!15, font=\ttfamily},
    activityRuns/.style = {base, minimum width=1cm, fill=orange!15, font=\ttfamily},
         process/.style = {base, minimum width=1cm, fill=orange!15, font=\ttfamily},
}
\def\BibTeX{{\rm B\kern-.05em{\sc i\kern-.025em b}\kern-.08em
    T\kern-.1667em\lower.7ex\hbox{E}\kern-.125emX}}
\begin{document}

\title{Online Causal Structure Learning in the Presence of Latent Variables\\
\thanks{}
}

\author{\IEEEauthorblockN{Durdane Kocacoban}
\IEEEauthorblockA{\textit{Department of Computer Science} \\
\textit{University of York}\\
York, UK \\
dk796@york.ac.uk}
\and
\IEEEauthorblockN{James Cussens}
\IEEEauthorblockA{\textit{Department of Computer Science} \\
\textit{University of York}\\
York, UK \\
james.cussens@york.ac.uk}
}

\maketitle

\begin{abstract}
We present two online causal structure learning algorithms which can track changes in a causal structure and process data in a dynamic real-time manner. Standard causal structure learning algorithms assume that causal structure does not change during the data collection process, but in real-world scenarios, it often does change. Therefore, it is inappropriate to handle such changes with existing batch-learning approaches, and instead, a structure should be learned in an online manner. The online causal structure learning algorithms we present here can revise correlation values without reprocessing the entire dataset and use an existing model to avoid relearning the causal links in the prior model, which still fit data. Proposed algorithms are tested on synthetic and real-world datasets. The online causal structure learning algorithms outperformed standard FCI by a large margin in learning the changed causal structure correctly and efficiently when latent variables were present.
\end{abstract}


\section{Introduction}
In this paper, we propose to use Causal Bayesian Networks (CBN), which play a central role in dealing with uncertainty in AI. Causal models can be created based on information, data, or both. Regardless of the source of information used to create the model, there may be inaccuracies, or the application area may vary. Therefore, the model needs constant improvement during use.

In the literature, there exist many kinds of causal structure learning algorithms which have been developed successfully and applied to many different areas \cite{pearl2000,SpirtesGlymourScheines2000,Chickering2002,KenettRaanan2010,Hyttinen2013}. Although all are successful structure learning algorithms, almost all these algorithms share an essential feature. They assume that causal structure does not change during the data collection process. In real-world scenarios, a causal structure often changes \cite{Kummerfeld2015}.  To quickly identify these changes and then learn a new structure are both crucial. Therefore, it is not possible to determine these changes with existing batch-learning approaches; instead, the structure must be learned in an online manner. 

We have a few online learning algorithms  \cite{Kummerfeld2012,Kummerfeld2013,Roure2004,Shi2010} in the literature, which are capable of detecting changes. However, none of them is capable of causal learning in the presence of latent and selection variables.

We present two heuristic algorithms which aim to fill these gaps. The first online causal structure algorithm we propose here is Online Fast Causal Inference (OFCI). OFCI is an online version of the Fast Causal Inference (FCI) algorithm. This algorithm is modified using the FCI instead of the PC algorithm in the DOCL algorithm of Kummerfeld and Danks \cite{Kummerfeld2012} to learn in the presence of hidden variables. The algorithm asserts that when given a learned causal structure, as new data points arrive the correlations will be revised with the estimation of the weight of each causal interaction, and the structure will be relearned when data present evidence that the underlying structure has changed. In particular, the method can estimate the causal structure even when its structure is changed multiple times and provides us with the learning structure at any time. 

The second online causal structure algorithm we propose here is Fast Online Fast Causal Inference (FOFCI). FOFCI is a modified version of OFCI in a way to minimise the learning cost of the current model by using the causal relationships between the variables of the previous model. Unlike OFCI, FOFCI not only updates the existing correlations in the light of new data but also uses the current causal structure in an attempt to speed up learning the new causal structure. FOFCI is faster than OFCI and FCI when some causal links in the current model still fit incoming data. When the current model is completely changed, the performance of FOFCI is identical to the OFCI.

Briefly, this paper is organised as follows. We begin with related work and then follow required definitions which allow us to understand algorithms. Next, we provide a detailed description of the online causal structure learning algorithms. Next, experimental results are given as evidence of successful learning of causal structure in the presence of confounding factors in a dynamic environment. In this study, we assume the data has a multivariate Gaussian distribution. 

We will evaluate the performance of algorithms under conditions where the causal structure is completely changed and slightly changed. First, the FCI and OFCI algorithms' performances are compared by learning performance and average running time when the causal structure is completely changed. We ascertain that OFCI outperforms FCI when the causal structure is changed. As the number of variables grows, OFCI substantially outperforms FCI in terms of computation time when the causal model is changing. 

Furthermore, we show that the learning performance of OFCI is identical to FCI when the causal structure does not change, so there is no penalty for being able to handle changing structure. It suggests that the online causal learning algorithms proposed here are practical alternative algorithms to batch learning algorithms. 

Next, we compared FCI, OFCI and FOFCI concerning learning performance and the number of independence tests applied when the causal model does not completely change. We note that the online causal structure learning algorithms are better at learning the true causal model than FCI for larger graphs. Also, we see that FOFCI learns the causal model using fewer independence tests than OFCI for all parameter settings. Last, we compare the average running time of OFCI and FOFCI, and FOFCI is faster than OFCI to learn the causal model. In a nutshell, we propose an efficient algorithm (FOFCI) with both low computational and sample complexity in contrast with OFCI. We terminate with a review of potential future developments of the given algorithm.
%%%%%%%%%%%%%%%%%%%%%%%%%%%%%%%%%%%%%%%%%%%%%%%%%%%%%%%%%%%%%%%%%%%%%%%%%%%%%%%%%%%%%%%%%%%%%%%%%%%%%%%%%%%%%%%%%%%%%%%%%%%%
%%%%%%%%%%%%%%%%%%%%%%%%%%%%%%%%%%%%%%%%%%%%%%%%%%%%%%%%%%%%%%%%%%%%%%%%%%%%%%%%%%%%%%%%%%%%%%%%%%%%%%%%%%%%%%%%%%%%%%%%%%%%
\section{Related works}
Bayesian network structure learning has attracted great attention in recent years. In a nutshell, structure learning can be defined as finding a DAG that fits the data. There are two distinct approaches to structure learning: score-based and constraint-based. In a score-based approach, each DAG gets a score based on how well it fits the data, and the goal is to find a DAG that maximises its score \cite{Cooper1990,Heckerman1995}. On the other hand, the constraint-based approach, also known as conditional independence test-based approaches, is based on testing conditional independence between variables and finding the best DAG that represents these relations \cite{Koller2009}. 

The critical point for the current research is the assumption that both of the causal structure learning approaches assume that the data comes from a single generating causal structure, and therefore these methods cannot be used directly for learning when a causal structure changes during the data collection process. Both types of approaches are not able to keep up with systems in a developing and changing world. Therefore, we need new tools to handle this, which are capable of giving results in a reasonable amount of time. Nevertheless, they only require sufficient statistics as input data and can, therefore, provide part of the solution to this problem. They need a mechanism which can detect changes, respond to it, and then learn the new causal model.

In the literature, there exist two main methods for online tracking of some feature in a structure, which are temporal difference learning (TDL) \cite{Sutton1988} and Bayesian change-point detection (CPD) \cite{AdamsMacKay2007}. Nevertheless, both methods have not been applied to detect changes in a casual structure, so they need some modifications to do this.  

The standard TDL algorithm provides a dynamic estimate of a univariate random variable using a simple update rule. In this update, the error in the current estimate is updated with a learning rate coefficient. Therefore, the static learning rate plays an important role and controls how quickly or slowly, a model learns a problem. If it is chosen too small, the TDL algorithm converges slowly. If it is chosen too large, the algorithm will be so sensitive even when the environment is stable. TDL algorithm can detect slow change but not high stability or dramatic changes. That feature is essential for causal structure learning as causal structures often have non-deterministic connections.

In contrast, CPD algorithms are useful for dramatic changes that indicate breaks between periods of stability \cite{AdamsMacKay2007}. CPD algorithms must store large parts of input data. These algorithms assume that the model only has a dramatically changing environment separated by periods of stability. Both algorithm types have not been applied to tracking causal structure. To do so, they need the necessary modifications. 

Talih and Hengartner \cite{TalihHengartner2005} do other related work. In their work, the data sets are taken sequentially as input and divided into a fixed number of data intervals, each with an associated undirected graph that differentiates one edge from its neighbours. In contrast to our work, in their work, they focus on a particular type of graphical structure change (a single edge added or removed), only work in batch mode and use undirected graphs instead of directed acyclic graphical models. Next, Siracusa \cite{Siracusa2009} uses a Bayesian approach to find posterior uncertainty on possible directed edges at different points in a time series. Our work differs from their work because we use frequentist methods instead of Bayesian methods, and we can work in real time in an incoming data stream.

Some methods aim to estimate the time-varying causal model. The DOCL algorithm proposed by Kummerfeld and Danks is applied to tracking causal structure \cite{Kummerfeld2012}. They demonstrated the adequate performance of algorithms in tracking changes in structure. It is important to note that we use their algorithm for change detection. Therefore, their work is the source of inspiration for our work.

On the other hand,  the work of Kummerfeld and Danks \cite{Kummerfeld2012} differs from ours in two ways. First of all, DOCL does not allow for the possibility of latent and selection variables and relearns a structure only at the points where the structure has changed. However, the critical problem with learning cause and effect from observational (as opposed to interventional) data is the presence of hidden confounders. Also, a causal structure can change multiple times throughout data collection. Next, DOCL runs the learning algorithm whenever there is a change in the structure. Also, the learned graph is used only to display the changing relationships between variables. However, in our algorithm, the relationships in the learned graphs take an active part in learning the next graph.

Another related method is proposed by Bendtsen \cite{Bendtsen2016}, which is the regime aware learning algorithm to learn a sequence of Bayesian networks that model a system with regime changes. These methods are not able to cope with real-world data as they suffer from a large number of statistical tests and ignore the existence of confounding factors. We present a new approach which is capable of detecting changes even multiple times and learning structure in the light of sequentially incoming data in the presence of confounding factors.
%%%%%%%%%%%%%%%%%%%%%%%%%%%%%%%%%%%%%%%%%%%%%%%%%%%%%%%%%%%%%%%%%%%%%%%%%%%%%%%%%%%%%%%%%%%%%%%%%%%%%%%%%%%
%%%%%%%%%%%%%%%%%%%%%%%%%%%%%%%%%%%%%%%%%%%%%%%%%%%%%%%%%%%%%%%%%%%%%%%%%%%%%%%%%%%%%%%%%%%%%%%%%%%%%%%%%%%
\section{Graphical Definitions}
A graph $G$ is a pair $(V, E)$ where $V$ is a set of variables, here corresponding to random variables, and $E$ is a set of edges. A directed acyclic graph (DAG) is a graph which has only directed edges and contains no directed cycles \cite{Heckerman1995}. A causal directed acyclic graph is a graph whose edges can be interpreted as causal relations. A causal DAG is \emph{causally sufficient}, if and only if every cause of two variables in the set is also in the set \cite{Koller2009}.

Three random vertices $\alpha$, $\beta$ and $\gamma$ are called an unshielded triple if $\alpha$ and $\beta$ are adjacent, $\beta$ and $\gamma$ are adjacent, but $\alpha$ and $\gamma$ are not adjacent. If two random vertices $\alpha$, $\beta$ are independent given a set $S$, then $S$ is called a separation set of $\alpha$ and $\beta$.  A maximal ancestral graph (MAG) is an ancestral graph where each missing edge corresponds to a conditional independence relationship \cite{colomborichardson2011}. If the observed variables of two DAGs encode all the same conditional independence relations, they are called Markov equivalent \cite{VermaPearl1991}. A partial ancestral graph (PAG) represents a Markov equivalence class of MAGs \cite{MalinskySpirtes2016}. The PAGs we will study can have (a subset of) the following edges: $\rightarrow$ (directed), $\leftrightarrow$ (bi-directed), $-$ (undirected), $\circcirc$ (nondirected), $\circline$ (partially undirected) and $\circrightarrow$ (partially directed). Although the FCI algorithm allows the possibility of latent and selection variables, selection variables are not considered in this study. In this study, we will not describe how algorithms deal with hidden variables because they are based on the FCI algorithm for causal model learning part \cite{SpirtesGlymourScheines2000}.
%%%%%%%%%%%%%%%%%%%%%%%%%%%%%%%%%%%%%%%%%%%%%%%%%%%%%%%%%%%%%%%%%%%%%%%%%%%%%%%%%%%%%%%%%%%%%%%%%%%%%%%%%%%%%%%%%%%%%%%%%%%%
%%%%%%%%%%%%%%%%%%%%%%%%%%%%%%%%%%%%%%%%%%%%%%%%%%%%%%%%%%%%%%%%%%%%%%%%%%%%%%%%%%%%%%%%%%%%%%%%%%%%%%%%%%%%%%%%%%%%%%%%%%%%
\subsection{FCI algorithm}
Spirtes \cite{spirtes1997} proposed the Fast Causal Inference (FCI) algorithm. FCI is a modified version of the PC algorithm, allowing arbitrarily many hidden and selection variables. It allows for the existence of hidden and selection variables and has been designed to show conditional independence and causal information between random variables \cite{colomborichardson2011}. The Oracle version of the FCI algorithm is shown in Algorithm~\ref{alg:fci} \cite{Colombo2012}.

\begin{algorithm}[H]
\footnotesize
\caption{FCI algorithm (oracle version)}
\label{alg:fci}
\textbf{Require:} Conditional independence information among all variables.
\begin{algorithmic}
\STATE Form a complete graph on the set of variables, where there is an edge $\circcirc$ between each variable pair.
\STATE Find the skeleton (an undirected version of output) in the light of independence tests and separations sets. 
\STATE Orient unshielded triples in the skeleton based on separation sets.
\STATE Use rules ($R1-R4$) and ($R8-R10$) from \cite{Zhang2008} to orient as many edges mark possible. 
\end{algorithmic}
\textbf{Return:} a $PAG$ which represent the conditional dependencies of the set of variables.
\end{algorithm}
%%%%%%%%%%%%%%%%%%%%%%%%%%%%%%%%%%%%%%%%%%%%%%%%%%%%%%%%%%%%%%%%%%%%%%%%%%%%%%%%%%%%%%%%%%%%%%%%%%%%%%%%%%%%%%%%%%%%%%%%%%%
%%%%%%%%%%%%%%%%%%%%%%%%%%%%%%%%%%%%%%%%%%%%%%%%%%%%%%%%%%%%%%%%%%%%%%%%%%%%%%%%%%%%%%%%%%%%%%%%%%%%%%%%%%%%%%%%%%%%%%%%%%%
\section{Problem Definition}
Given a set of continuous variables $V$, we assume that we have a true underlying causal model over $V$ at each moment in time. We specify a causal model by a pair  $\langle G, F\rangle$, where $G$ denotes a DAG over $V$, and $F$ is a set of linear equations. These kinds of causal models are also known as recursive Causal Structural Equation Models (SEMs) \cite{Kummerfeld2012}. We assume that the data are independently generated from the true underlying causal model at each moment in time, though we do not assume that this causal model is stationary through time.

In a nutshell, the online causal structure learning algorithms proposed here take a new datapoint as input at each time step and outputs a graphical model (PAG). 

More precisely, the algorithms are separated into three functionally different parts; 

First, the Online Covariance Matrix Estimator (OCME) \cite{Kummerfeld2012} takes each datapoint sequentially as input and updates the sufficient statistics (covariance matrix, sample size and mean) \cite{Kummerfeld2013,Kummerfeld2012}. In particular, OCME maintains an estimated covariance matrix over the variables and updates the estimated covariance matrix in return for incoming datapoints. As OCME does not store any of the incoming new datapoints, its memory needs only the estimated covariance matrix. In contrast, batch mode algorithms require the memory both all data-samples and the estimated covariance matrix \cite{Kummerfeld2013,Kummerfeld2012}. Thus, the proposed algorithms have a substantial memory advantage compared to batch mode learning algorithms. As we do not assume a stationary causal model, the datapoints should be weighted differently in a way to weight more recent datapoints more heavily after a change occurs and reduce confidence in previous data points. 

Any causal structure learning algorithm needs a sample size with an estimated covariance matrix. In the proposed algorithms, we also need to update the sample size. We assume that every new data point contributes 1 to the sample size. However, different datapoints can get different weights, and the sample size should be updated accordingly. Every new datapoint should get more or equal weight than previous datapoints. Therefore, sample size, which is called an effective sample size \cite{Kummerfeld2012}, does not necessarily have to be equal to true sample size (which should be less than or equal to the actual sample size). 

Next, the Causal Model Change Detector (CMCD) \cite{Kummerfeld2012} tracks the fitness between the current estimated covariance matrix and the input data to detect the changes in the underlying causal model.  It requires adjusting to the previous and new datapoints' relative weight \cite{Kummerfeld2013,Kummerfeld2012}. Specifically, the fit between each incoming datapoint and the current estimated covariance matrix is given by the Mahalanobis distance. A large Mahalanobis distance for any particular datapoint can merely indicate an outlier; consistently large Mahalanobis distances over multiple datapoints state that the current estimated covariance matrix fits poorly to the underlying causal model. Therefore, the new datapoints should be weighted more heavily \cite{Kummerfeld2013,Kummerfeld2012}. The approach is to first calculate the individual $p$-values for each datapoint. The Mahalanobis distance of a $V$-dimensional datapoint from a covariance matrix estimated from a sample of size is distributed as Hotelling's $T^2$. So then, a weighted pooling method to aggregate those $p$-values into a pooled $p$-value by using Liptak's method \cite{Liptak1958} is used. Finally, the weight of the next point, given the pooled $p$-value, is determined. 

The Causal Model Learner (CML) \cite{Kummerfeld2012} learns the causal model from the estimated (from weighted data) sufficient statistics (covariances, sample size and means) provided in OCME. Kummerfeld and Danks's \cite{Kummerfeld2012} algorithm uses the PC algorithm \cite{SpirtesGlymourScheines2000} as a standard constraint-based causal structure learning algorithm. 

In contrast with the method of Kummerfeld and Danks's work, OFCI uses FCI instead of PC. Just like Kummerfeld and Danks's work, OFCI relearns the causal model after a change occurs. Also, unlike the DOCL \cite{Kummerfeld2012}, we used the actual sample size for structure learning algorithms, but it is limited to just CML part. The effective sample size is used for OCME and CMCD.

Learning graphical model structure is computationally expensive, and so one should balance the accuracy of the current model against the computational cost of relearning. For this reason, we propose FOFCI to reduce the computational cost of relearning and make the proposed algorithms more online. The FOFCI algorithm differs from both OFCI and Kummerfeld and Danks's work \cite{Kummerfeld2012} for causal model learning the part. In those two algorithms, the OCME and CMCD parts continue to update sufficient statistics as long as only the new data point is available. However, the CML part has no other role than to learn the updated information. Nevertheless, in order to speak of a real online learning mechanism, all parts of the algorithm must be actively involved in learning at the next point of change. 

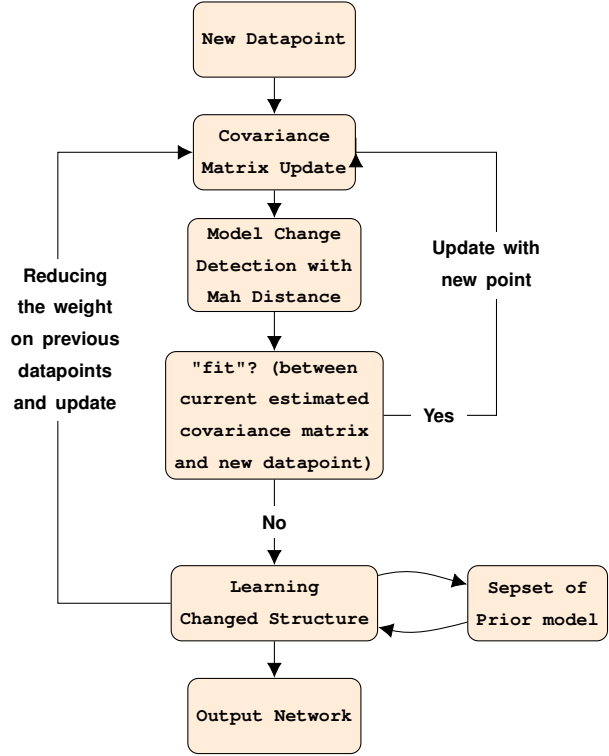
\begin{figure}
\centering
\caption{Basic flowchart of FOFCI algorithm} \label{fig:FOFCI}
 \begin{tikzpicture}[node distance=1.5cm,
    every node/.style={fill=white, font=\sffamily}, align=center][H]

  \node (start)             [activityStarts]              {\scriptsize \textbf{New Datapoint}};
  \node (onCreateBlock)     [process, below of=start]          {\scriptsize \textbf{Covariance}\\\scriptsize \textbf{Matrix Update}};
  \node (onStartBlock)      [process, below of=onCreateBlock]   {\scriptsize \textbf{Model Change}\\\scriptsize \textbf{Detection with}\\\scriptsize \textbf{Mah Distance}};
  \node (onResumeBlock) [process,below of=onStartBlock, yshift=-0.5cm]   {\scriptsize \textbf{"fit"? (between}\\ \scriptsize \textbf{current estimated}\\ \scriptsize \textbf{covariance matrix}\\ \scriptsize \textbf{and new datapoint)}};
  \node (activityRuns)      [activityRuns, below of=onResumeBlock, yshift=-1cm]
                                {\scriptsize \textbf{Learning}\\\scriptsize \textbf{Changed Structure}};
 \node (sepset)      [process, right of=activityRuns, xshift=2cm]
                                                        {\scriptsize \textbf{Sepset of}\\\scriptsize \textbf{Prior model}};
  \node (onPauseBlock)      [process, below of=activityRuns]
                                                                {\scriptsize \textbf{Output Network}};
  
%------------------------------------------------------------------------------
\draw[->]             (start) -- (onCreateBlock);
\draw[->]     (onCreateBlock) -- (onStartBlock);
\draw[->]      (onStartBlock) -- (onResumeBlock);
\draw[->]     (onResumeBlock) -- node[text width=4cm]
                                   {\scriptsize \textbf{No}} (activityRuns);
\draw[->]      (activityRuns) -- (onPauseBlock);
\draw [->] (activityRuns) to [out=15,in=165] (sepset);
\draw [->] (sepset) to [out=195,in=-13] (activityRuns);
\draw[->] (onResumeBlock.east) -- ++(1.5,0) -- ++(0,1.5) -- ++(0,2) --                
    node[xshift=0.8cm,yshift=-1.5cm, text width=1.5cm]
   {\scriptsize \textbf{Update with new point}}(onCreateBlock.east)--node[xshift=1.1cm,yshift=-3.5cm, text width=0.5cm]
   {\scriptsize \textbf{Yes}}(onCreateBlock.east);
\draw[->] (activityRuns.west) -- ++(-1.5,0) -- ++(0,3.5) -- ++(0,2.5) --                
    node[xshift=-0.8cm,yshift=-2.5cm, text width=1.5cm]
   {\scriptsize \textbf{Reducing the weight on previous datapoints and update}}(onCreateBlock.west);
\end{tikzpicture}
\end{figure}

In contrast, FOFCI uses a modified version of the FCI algorithm. In this modified version, the algorithm takes the separation sets of the previous model as input, unlike the classic FCI. In the first part of the algorithm, it is found out whether the causal links of the previous model's separation sets still fit the updated covariance matrix.  If some of them still fit, the independence tests that will be applied to determine these relations are eliminated. As can be seen in the experimental results section, this sometimes reduces the independence test by fifty percent. Thus, it saves us from the unnecessary test repetition that can find thousands for large networks. This allows us to start analysing on a more straightforward graph rather than starting from a complete graph like in the classic FCI. The rest of the algorithm continues the same as in the classic FCI. This simple Figure~\ref{fig:FOFCI} represents a process of FOFCI.

In particular, as it is seen in the Figure~\ref{fig:FOFCI}, OCME first updates the estimated covariance matrix in response to incoming datapoints, CMCD tracks the fitness between the current estimated covariance matrix and the input data. Unlike Kummerfeld and Danks's work and OFCI, CML takes the covariance matrix, and also separation sets of the previous learned casual model as input. Structure learning part starts with checking of causal links in separation set in the prior model. If all or some causal links of the prior model still fits incoming data, we do not need to apply the independence tests which are required to find these causal links.  

Then, the structure learning algorithm finds the initial skeleton by starting with the graph obtained after this analysis and updates the separation sets at the same time. After learning the causal model, the separation sets which are updated according to the new model are stored to use in the next change point. The process of FOFCI will be identical to OFCI in cases where the causal structure is completely changed. By comparing to OFCI and DOCL, FOFCI seems to need more memory space to store separation sets of learned models, but it performs significantly better than two algorithms in terms of time and space complexity. 

The relearning should be most frequent after an inferred underlying change, though there should be a non-zero chance of relearning even when the structure appears to be relatively stable. Kummerfeld's work and OFCI have the limitations such as the over the computational cost of relearning of stable parts in cases where only some parts of the causal structure are changed.

Therefore, FOFCI fills this gap. Optionally, a probabilistic relearning scheduler is added to the algorithm, which utilises the pooled p-values calculated in the CMCD module to determine when to relearn the causal model. 

For the first two parts (OCME and CMCD), in this study, we just described these parts and their functions. OCME and CMCD are identical in DOCL, OFCI and FOFCI and belong to Kummerfeld and Danks. Therefore, in-depth mathematical knowledge (equations, properties, theorem and proof) of these parts can be found in Kummerfeld and Danks's works \cite{Kummerfeld2012,Kummerfeld2013}.
%%%%%%%%%%%%%%%%%%%%%%%%%%%%%%%%%%%%%%%%%%%%%%%%%%%%%%%%%%%%%%%%%%%%%%%%%%%%%%%%%%%%%%%%%%%%%%%%%%%%%%%%%%%%%%%%%%%%%%%%%%%
%%%%%%%%%%%%%%%%%%%%%%%%%%%%%%%%%%%%%%%%%%%%%%%%%%%%%%%%%%%%%%%%%%%%%%%%%%%%%%%%%%%%%%%%%%%%%%%%%%%%%%%%%%%%%%%%%%%%%%%%%%%
\section{Experimental Results}
We evaluate the performance of the FCI, OFCI and FOFCI algorithms on synthetic and real data. The results confirm the effectiveness of the proposed algorithms.
%%%%%%%%%%%%%%%%%%%%%%%%%%%%%%%%%%%%%%%%%%%%%%%%%%%%%%%%%%%%%%%%%%%%%%%%%%%%%%%%%%%%%%%%%%%%%%%%%%%%%%%%%%%%%%%%%%%%%%%%%%%
%%%%%%%%%%%%%%%%%%%%%%%%%%%%%%%%%%%%%%%%%%%%%%%%%%%%%%%%%%%%%%%%%%%%%%%%%%%%%%%%%%%%%%%%%%%%%%%%%%%%%%%%%%%%%%%%%%%%%%%%%%%
\subsection{Application to Synthetic Datasets}

Synthetic datasets are used to verify the accuracy of our online algorithms inference approach when given a known ground truth network. Results are evaluated under the condition where the true partial ancestral graph is changed during the data collection process. 

We have created each synthetic dataset by following the same procedure by using the \textbf{pcalg} package for R \cite{KalishMarcus2012}. First, we generated four random DAGs, which each have the same number of nodes, and all are different from each other. For each DAG, we use the following procedure. Each random DAG is generated with a given number of vertices $p' $, expected neighbourhood size $E(N)$ and sample size 10000.  Next, we concatenated the data from these four different graphs that have the same characteristics (vertices, $E(N)$ and sample size) to obtain a dataset with 40000 samples. Therefore the dataset is created by aggregating four different graphs' distributions. That means there are three change points in each data. 

We do this to see the performance of OFCI and FOFCI in the case where the causal structure is changed multiple times. We restrict each graph to have two latent variables that have no parents and at least two children. (Selection variables are not considered in this study).

Our goal is to present two algorithms that work in real-world scenarios, that are not only tracking the change of causal structure but also can compete with existing batch structure learning algorithms concerning cost even when the causal structure does not change. 

The algorithms would relearn the causal model for each time when the data presents evidence that the underlying causal structure has changed and output a PAG. The algorithm will continue to update as long as only the new data point is available. For this work, examining each learned model for a data set which has 40000 sample size will not be possible. Since we already know the main structural change points in the data set, we have just added a relearning scheduler to the algorithm to see the performance at these points. The experimental results are divided into two parts. 

For the first part, FCI and OFCI are compared to display the performance of the online algorithm when the structure is completely changed. For each value of $p'\in$ \{8, 10, 13, 15, 18\}, we generated 40 random DAGs with $E(N)=$2. For each such DAG, we generated a data set of size $n=$10000 and ran FCI and OFCI with parameter $\alpha=$ 0.05. We used a small number of variables because it will be impossible to generate completely different graphs as the number of nodes grows. Therefore, more extensive graphs will be included in the second part.

The resulting evaluation of the desired algorithm is based on the exact true graph, which is a PAG used for generating dataset rather than the Markov equivalence class of the true graph. In general, the studies that represent the learning capabilities of such algorithms are based on the Markov equivalence class of the true causal model for comparison. In this part, we especially want to display the ability of the desired algorithm to detect true edge orientation. 

\begin{figure}[H]
\centerline{\includegraphics[width=8cm,height=4cm]{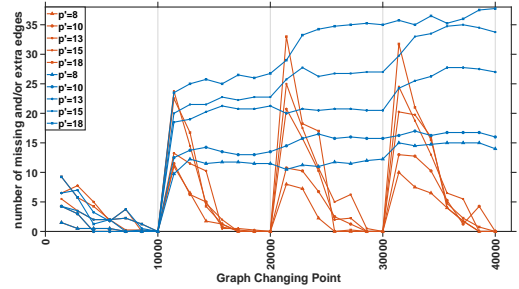}} 
\caption{FCI (blue) and OFCI (red), Average number of missing and/or extra edges}
\label{fig:three}
\end{figure}
Fig~\ref{fig:three} presents the average number of missing and/or extra edges when the causal structure changes 3 times at 10000, 20000 and 30000 during the data collection process. Zero means that there are no missing or extra edges and the algorithm works correctly. High numbers represent the poor fit to the true causal model. 

\begin{figure}[H]
\centerline{\includegraphics[width=8cm,height=4cm]{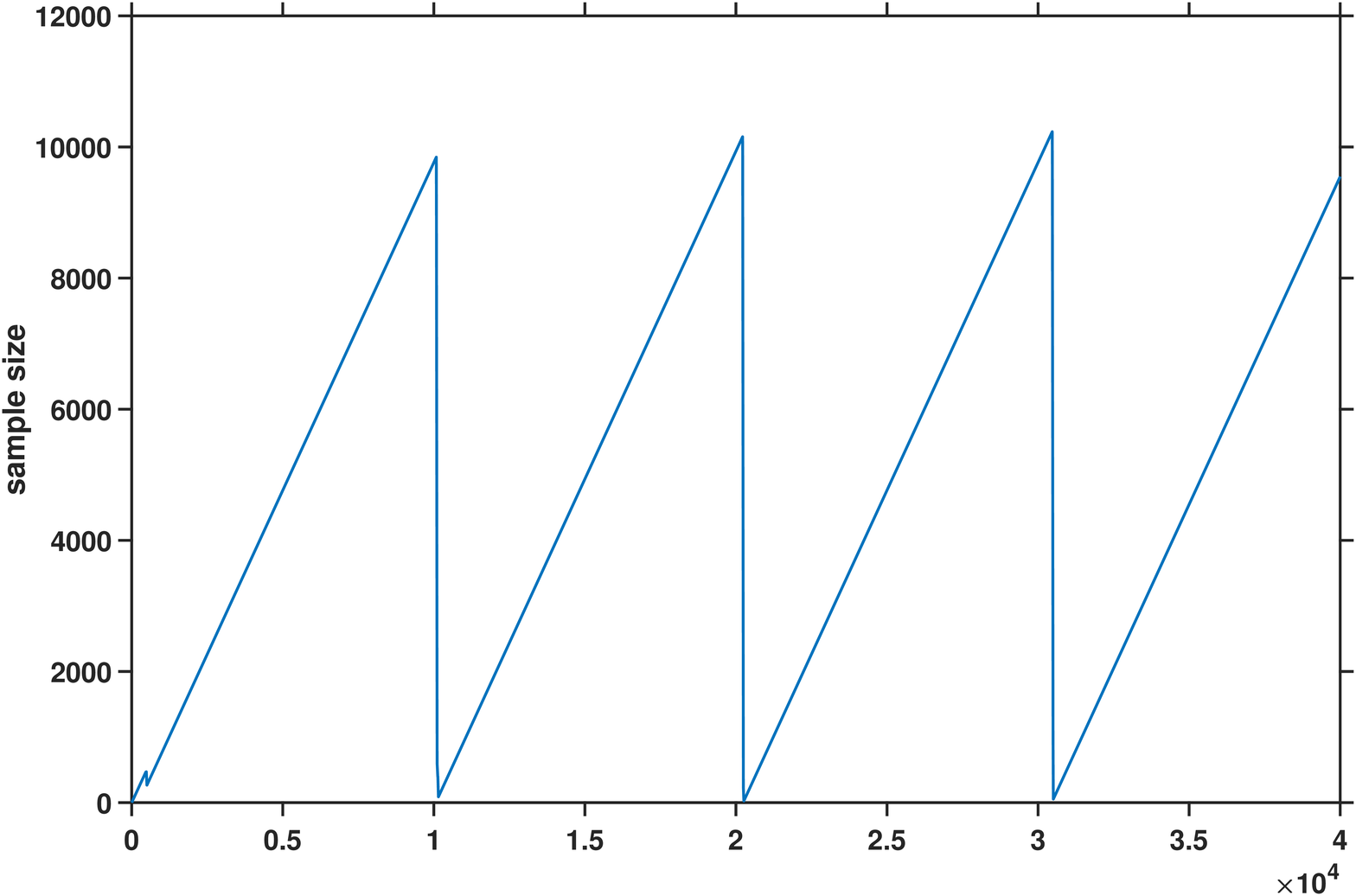}} 
\caption{Effective sample size}
\label{fig:five}
\end{figure}
\begin{figure}[H]
\centerline{\includegraphics[width=8cm,height=4cm]{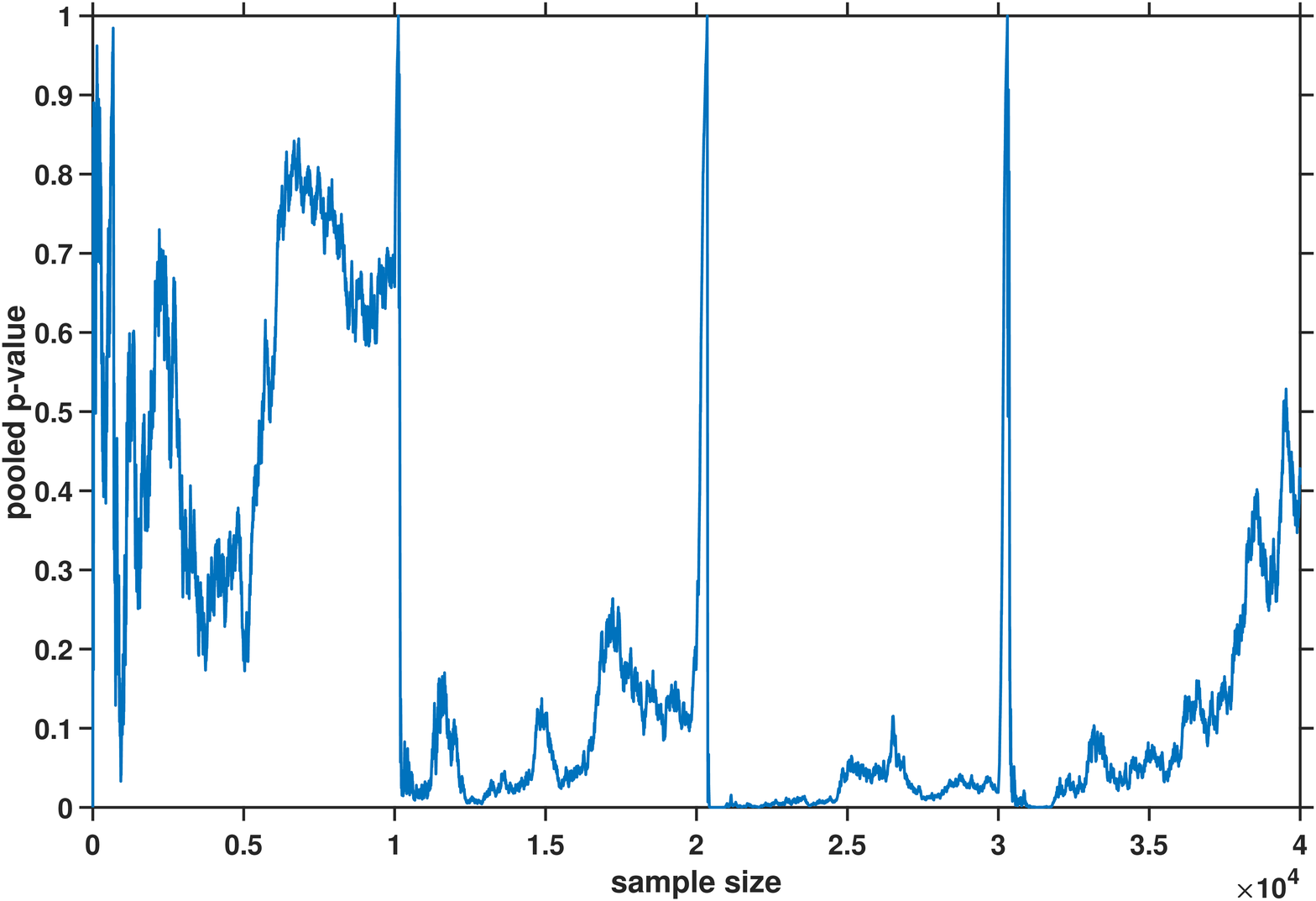}} 
\caption{Pooled p-values}
\label{fig:six}
\end{figure}
\begin{figure}[H]
\centerline{\includegraphics[width=8cm,height=4cm]{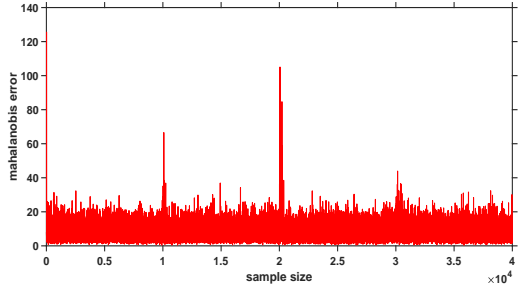}} 
\caption{Mahalanobis distances}
\label{fig:seven}
\end{figure}
\vspace{-1mm}
In Figs~\ref{fig:five}--\ref{fig:seven}, we present, respectively, the changing of sample size, pooled p-values and Mahalanobis distances during the learning process. We just used 18 variables, 40000 sample size example.

As is clear from the graphs, the performance of OFCI and FCI could not be distinguished from each other before datapoint 10000, which is the first changing point of the causal structure. As previously mentioned, a relearning scheduler is added to the desired algorithm (optionally). In this way, we scheduled learning process for 500, 1000, 3500, 5000, 7500, 9000, 10000, 11000, 13000, 15000, 17500, 18000, 19000, 20000, 20500, 23000, 25000, 27500, 28000, 29000, 30000, 30500, 33000, 35000, 37500, 38000, 39000, 40000 datapoints for OFCI. OFCI relearned the causal structure, and FCI was rerun after these datapoints. After the underlying causal structure is changed at 10000, 20000 and 30000 datapoints, OFCI significantly outperformed FCI. As the datasets are a mix of four different distributions that indicate a large number of synthetic variables, FCI works poorly after the first changing datapoint. OFCI measures major Mahalanobis distance for changing datapoints as can be seen from the example in Figure~\ref{fig:seven}. Therefore, it leads to higher weights and learns the new underlying causal structure.

\begin{table}[H]
\caption{Average Running Time in seconds for 20000, 30000 and 40000 sample size dataset during learning the changing structure.}
\label{tab:one}
\begin{center}
\begin{footnotesize}
\begin{sc}
\begin{tabular}{|lccr|}
\hline
Dataset ($|Vars|$)&\multicolumn{3}{c|}{20000}\\
& FCI & OFCI & Better?\\
\hline
 Dataset (8) &  \textbf{1.5} & 4.5 & $\times$ \\
 Dataset (10) & 4.5 &  \textbf{4.5} & \textbf{$\surd$} \\
 Dataset (13) & 27.5 &  \textbf{6} & \textbf{$\surd$} \\
 Dataset (15) & 57 &  \textbf{16} & \textbf{$\surd$} \\
 Dataset (18) & 259 &  \textbf{10.5} & \textbf{$\surd$} \\
\hline
 &\multicolumn{3}{c|}{30000}\\
&FCI & OFCI & Better?\\
\hline
 Dataset (8) &  \textbf{4.5} & 6.5 & $\times$ \\
 Dataset (10) &  13.5 &  \textbf{7} & \textbf{$\surd$} \\
 Dataset (13) &  74 &  \textbf{14.5} & \textbf{$\surd$} \\
 Dataset (15) &  225.5 &  \textbf{36} & \textbf{$\surd$} \\
 Dataset (18) &  1093.5 &  \textbf{94.5} & \textbf{$\surd$} \\
\hline
  &\multicolumn{3}{c|}{40000}\\
& FCI & OFCI& Better?\\
\hline
 Dataset (8) &  \textbf{7} & 9.5 & $\times$\\
 Dataset (10) & 23 &  \textbf{10} & \textbf{$\surd$} \\
 Dataset (13) & 171.5 &  \textbf{22.5} & \textbf{$\surd$} \\
 Dataset (15) & 542.5 &  \textbf{51.5} & \textbf{$\surd$} \\
 Dataset (18) & 2475 &  \textbf{202} & \textbf{$\surd$} \\
\hline
\end{tabular}
\end{sc}
\end{footnotesize}
\end{center}
\end{table}

The algorithm does not store any of datapoints coming sequentially. Its memory requirements are just for the estimated covariance matrix and sample size. Therefore, the algorithm has significant storage advantages for computational devices that cannot store all data. Additionally, online algorithms have an essential advantage concerning the computational time for complex networks.

We compare structure learning time differences (in seconds) between OFCI and FCI in Table~\ref{tab:one}. The Table~\ref{tab:one} represents learning processing time (in seconds) for 40 different datasets having sample sizes: 20000, 30000 and 40000. For datasets with just dataset(8), the FCI  outperforms OFCI by a small amount. However, for complex networks, OFCI outperforms FCI significantly because OFCI updates just the estimated covariance matrix. Especially, as the size of datasets grows, the performance gap becomes greater. 

Next, we investigated the performances of FCI, OFCI and our adaptation FOFCI, considering the number of differences in the output by comparing to the Markov equivalence class of the true DAG. As we made a large scale schedule review in the first part, in this part, the relearning scheduler is scheduled for just the main change points which are 10000, 20000, 30000, 40000 for OFCI and FOFCI. OFCI and FOFCI relearn the causal structure, and FCI was rerun after these datapoints. We used two simulation settings: small-scale and large-scale.

The small-scale simulation setting is as follows. For each value of $p'\in$ \{25, 30, 35, 40\}, we generated 160 random DAGs with $E(N)=$2. For each such DAG, we generated a data set that has  $n=$10000 sample size and ran FCI, OFCI, FOFCI with the $p$-value for independence tests set to $\alpha=$ 0.05.

The large-scale simulation setting is as follows. For each value of $p\in$ \{100, 125, 175, 200\} we generated 160 random DAGs with $E(N)=$2. For each DAG, we generated a data set of $n=$10000, and ran FCI, OFCI, FOFCI using parameter $\alpha=$ 0.05. In real-world datasets, some edges or adjacencies in the model may stay stable during data collection and learning process. Therefore, in this part, while generating random PAGs, we focused on small differences between structures to show the effectiveness of FOFCI where the causal structure does not dramatically change. 

\begin{figure}[H]
\begin{center}
\centerline{\includegraphics[width=8cm,height=4cm]{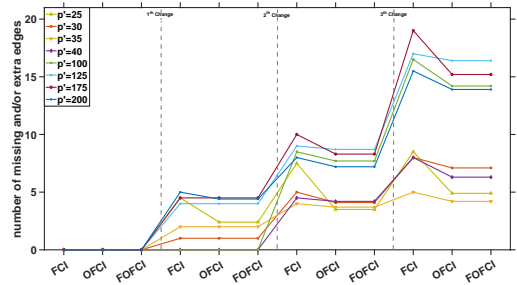}} 
\caption{FCI, OFCI and FOFCI, average number of miss and/or extra edges}
\label{fig:eight}
\end{center}
\end{figure}

Figure~\ref{fig:eight} shows the results for the small and large scale settings together. Figure~\ref{fig:eight} shows the average number of missing or extra edges over 5 replicates, and we see that this number was virtually identical for all algorithms. As expected, OFCI and FOFCI perform the same to learn the true causal model. We note that OFCI and FOFCI also outperform to FCI for learning the causal model for more massive graphs. Zero means that there are no missing or extra edges, and the algorithm works correctly. High numbers represent the poor fit to the true causal model. 

\begin{figure}[H]
\centerline{\includegraphics[width=8cm,height=4cm]{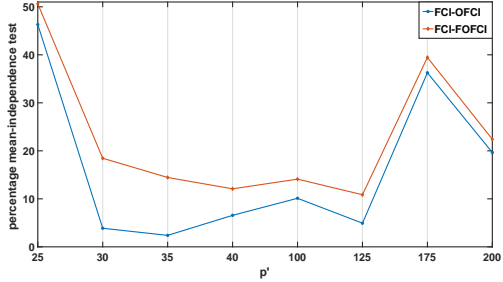}} 
\caption{OFCI and FOFCI, average percentage reduction of independence test in contrast with FCI}
\label{fig:nine}
\end{figure}
\begin{figure}[H]
\centerline{\includegraphics[width=8cm,height=4cm]{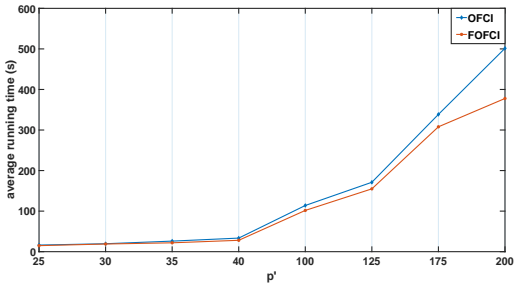}} 
\caption{ OFCI and FOFCI, average running time for learning structure}
\label{fig:ten}
\end{figure}

Next, we investigated the number of independence tests of both online algorithms. In particular, we first determined the number of necessary independence tests to learn the causal model for FCI, OFCI and FOFCI. Next, we stated the percentage difference of independence test numbers as FCI-OFCI and FCI-FOFCI. Figure~\ref{fig:nine} shows the average percentage difference of independence test number in small and large-scale settings. We see that FOFCI requires fewer independence tests compared to OFCI to learn the causal model for all the same parameter settings. High numbers represent the success of the algorithm in pruning the search space. We continued with a comparison of the learning time of the sample version of OFCI and FOFCI under the same simulation settings.

Figure~\ref{fig:ten} shows the average running times in the small and large-scale setting. We see that FOFCI is faster for all parameter settings. FOFCI learned the causal models faster than OFCI. As the scale expands, the difference between them also grows can be seen in detail in Table~\ref{tab2}. 

\begin{table}[H]
\centering
\caption{Average Running Time in seconds for 40000 sample size data}\label{tab2}
\begin{tabular}{|l|c|c|c|}
\hline
Dataset ($|Vars|$) & OFCI & FOFCI & Better? \\
\hline
 Dataset (25) & 16.3025 & {\bfseries 14.8766} & $\surd$ \\
 Dataset (30) & 19.5713 & {\bfseries 18.9014} & $\surd$ \\
 Dataset (35) & 26.0471 & {\bfseries 21.8677} & $\surd$ \\
 Dataset (40) & 33.4834 & {\bfseries 28.041} & $\surd$ \\
 Dataset (100) & 113.8983 & {\bfseries 101.6976} & $\surd$ \\
 Dataset (125) & 171.1545 & {\bfseries 154.8938} & $\surd$ \\
 Dataset (175) & 338.5137 & {\bfseries 307.9159} & $\surd$ \\
 Dataset (200) & 501.4344 & {\bfseries 377.7653} & $\surd$ \\
\hline
\end{tabular}
\end{table}
%%%%%%%%%%%%%%%%%%%%%%%%%%%%%%%%%%%%%%%%%%%%%%%%%%%%%%%%%%%%%%%%%%%%%%%%%%%%%%%%%%%%%%%%%%%%%%%%%%%%%%%%%%%%%%%%%%%%%%%%%%%
%%%%%%%%%%%%%%%%%%%%%%%%%%%%%%%%%%%%%%%%%%%%%%%%%%%%%%%%%%%%%%%%%%%%%%%%%%%%%%%%%%%%%%%%%%%%%%%%%%%%%%%%%%%%%%%%%%%%%%%%%%%
\subsection{Application to a Real-World Dataset}
We have applied the online algorithm to seasonally adjusted price index data available online from the U.S. Bureau of Labor Statistics to confirm the efficiency of the change detection part of the online learning algorithms (so this part is identical in both algorithms). We have limited the data to commodities extending to at least 1967 and resulting in a data set of 6 variants: Apparel, Food, Housing, Medical, Other, and Transportation. Data were collected monthly from 1967 to 2018 and reached 619 data points. Due to significant trends in the indices over time, we used the month-to-month differences.

\begin{figure}[H]
\centerline{\includegraphics[width=8cm,height=4cm]{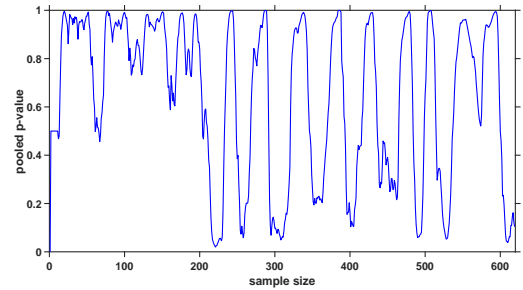}} 
\caption{Pooled p-values}
\label{fig:eleven}
\end{figure}
\begin{figure}[H]
\centerline{\includegraphics[width=8cm,height=5.5cm]{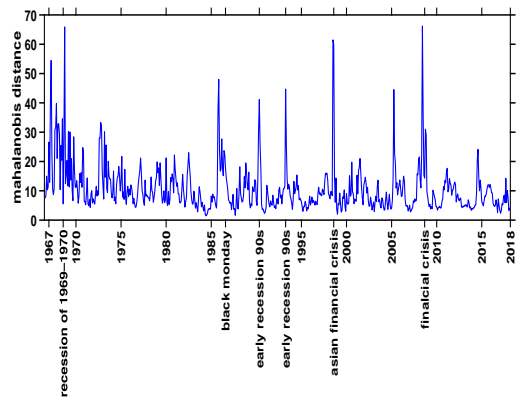}} 
\caption{Mahalanobis distances}
\label{fig:twelwe}
\end{figure}

Figure~\ref{fig:eleven} and Figure~\ref{fig:twelwe} show the drivers of these changes: the pooled p-value and Mahalanobis distance that has been collected for each month. Notably, the proposed algorithm detects a shift in the volatility of the causal relationships among these price indexes around recession of 1969-1970, the black Monday in 1987, 1990s early recession, Asian financial crisis in 1997 and Global financial crisis in 2007-2008. This real-world case study also shows the importance of using pooled p-values. As this study aims to propose an efficient algorithm for real-world cases, we will not measure financial performance; analysing and interpreting on this data.

The simulations were performed on a dual-core Intel Core i5 with 2.6 GHz and 16 GB RAM on macOS using Matlab R2018a.
%%%%%%%%%%%%%%%%%%%%%%%%%%%%%%%%%%%%%%%%%%%%%%%%%%%%%%%%%%%%%%%%%%%%%%%%%%%%%%%%%%%%%%%%%%%%%%%%%%%%%%%%%%%%%%%%%%%%%%%%%%%%
%%%%%%%%%%%%%%%%%%%%%%%%%%%%%%%%%%%%%%%%%%%%%%%%%%%%%%%%%%%%%%%%%%%%%%%%%%%%%%%%%%%%%%%%%%%%%%%%%%%%%%%%%%%%%%%%%%%%%%%%%%%%
\section{Discussion and future research}
In this paper, we introduce the OFCI (Kummerfeld and Danks's work-PC+FCI) and a new algorithm called FOFCI for learning PAGs. We evaluated the performance of these algorithms by testing them on synthetic and real data. The results show the efficacy of the proposed algorithms compared to FCI.

The outputs of OFCI and FOFCI are identical to each other and better than FCI. Also, FOFCI requires fewer conditional independence tests than OFCI and FCI to learn the causal model for both small and large numbers of variables. Additionally, we showed that FOFCI is faster than OFCI due to the smaller search space of the FOFCI algorithm. Therefore, we can say that FOFCI is the most efficient algorithm in this study.

Both online algorithms are useful for learning changing causal structure. We showed that the algorithms could track changes and learn new causal structure in a reasonable amount of time. However, the algorithms have limitations. Sometimes, the new model learning process of algorithms takes a long time because they require most of the data samples to learn the true model. Therefore, this means that online algorithms will perform poorly if the causal structure changes rapidly.

OFCI and FOFCI have a plug-and-play design. This feature allows for easy modification to use alternative algorithms. A range of alternative structure learning algorithms could be used instead of FCI, constraint-based methods such as RFCI \cite{KalishMarcus2012} and score based methods such as greedy search algorithms, depending on the assumptions one can make. Thus, the developments in structure learning algorithms will automatically improve the performance of this online structure learning algorithm. 

OFCI and FOFCI can track sufficient statistics for a linear Gaussian system efficiently. This problem is much harder for categorical/discrete variables or non-linear systems, as there will typically not be any compact representation of the sufficient statistics. One potential advantage of this approach is a way to learn conditional independence constraints in an online fashion, and then those constraints can be fed into any structure learning algorithm we want.
\section*{Acknowledgements} 
We want to thank Erich Kummerfeld, David Danks and David Heckerman warmly for valuable recommendations, helps and sharing their sources with us.
\bibliography{example_paper.bib}
\bibliographystyle{IEEEtran}
\end{document}